\documentclass[10pt,twocolumn,letterpaper]{article}

\usepackage{cvpr}
\usepackage{times}
\usepackage{epsfig}
\usepackage{graphicx}
\usepackage{amsmath}
\usepackage{amssymb}
\usepackage{subfigure}
\usepackage{diagbox} 
\usepackage{tabularx}
\usepackage{url}
\usepackage{verbatim}
\usepackage{lipsum}

\usepackage[breaklinks=true,letterpaper=true,colorlinks,bookmarks=false]{hyperref}

\cvprfinalcopy 

\def\cvprPaperID{****} 

\ifcvprfinal\fi
\begin{document}

\title{FaceX-Zoo: A PyTorch Toolbox for Face Recognition}

\author{Jun Wang, Yinglu Liu, Yibo Hu, Hailin Shi and Tao Mei\\
JD AI Research, Beijing, China\\
{\tt\small \{wangjun492, liuyinglu1, huyibo6, shihailin, tmei\}@jd.com}}

\maketitle
\newcommand\blfootnote[1]{%
\begingroup
\renewcommand\thefootnote{}\footnote{#1}%
\addtocounter{footnote}{-1}%
\endgroup
}

\begin{abstract}
$\blfootnote{
\textbf{Acknowledgements} Thanks to Jixuan Xu (\emph{jixuanxu@shu.edu. cn}), Hang Du (\emph{duhang@shu.edu.cn}), Haoran Jiang (\emph{jianghaoran@shu.edu.cn}) and Hanbin Dai  (\emph{daihanbin@std.uestc.edu.cn}), partial works were done when they interned at JD AI Research.}$
Deep learning based face recognition has achieved significant progress in recent years. Yet, the practical model production and further research of deep face recognition are in great need of corresponding public support.
For example, the production of face representation network desires a modular training scheme to consider the proper choice from various candidates of state-of-the-art backbone and training supervision subject to the real-world face recognition demand; for performance analysis and comparison, the standard and automatic evaluation with a bunch of models on multiple benchmarks will be a desired tool as well; besides, a public groundwork is welcomed for deploying the face recognition in the shape of holistic pipeline.
Furthermore, there are some newly-emerged challenges, such as the masked face recognition caused by the recent world-wide COVID-19 pandemic, which draws increasing attention in practical applications. A feasible and elegant solution is to build an easy-to-use unified framework to meet the above demands.
To this end, we introduce a novel open-source framework, named FaceX-Zoo, which is oriented to the research-development community of face recognition.
Resorting to the highly modular and scalable design, FaceX-Zoo provides a training module with various supervisory heads and backbones towards state-of-the-art face recognition, as well as a standardized evaluation module which enables to evaluate the models in most of the popular benchmarks just by editing a simple configuration.
Also, a simple yet fully functional face SDK is provided for the validation and primary application of the trained models.
Rather than including as many as possible of the prior techniques, we enable FaceX-Zoo to easily upgrade and extend along with the development of face related domains.
The source code and models are available at: \href{https://github.com/JDAI-CV/faceX-Zoo}{https://github.com/JDAI-CV/FaceX-Zoo}.
\end{abstract}


\begin{figure*}[h]
    \centering
    \includegraphics[width=\linewidth]{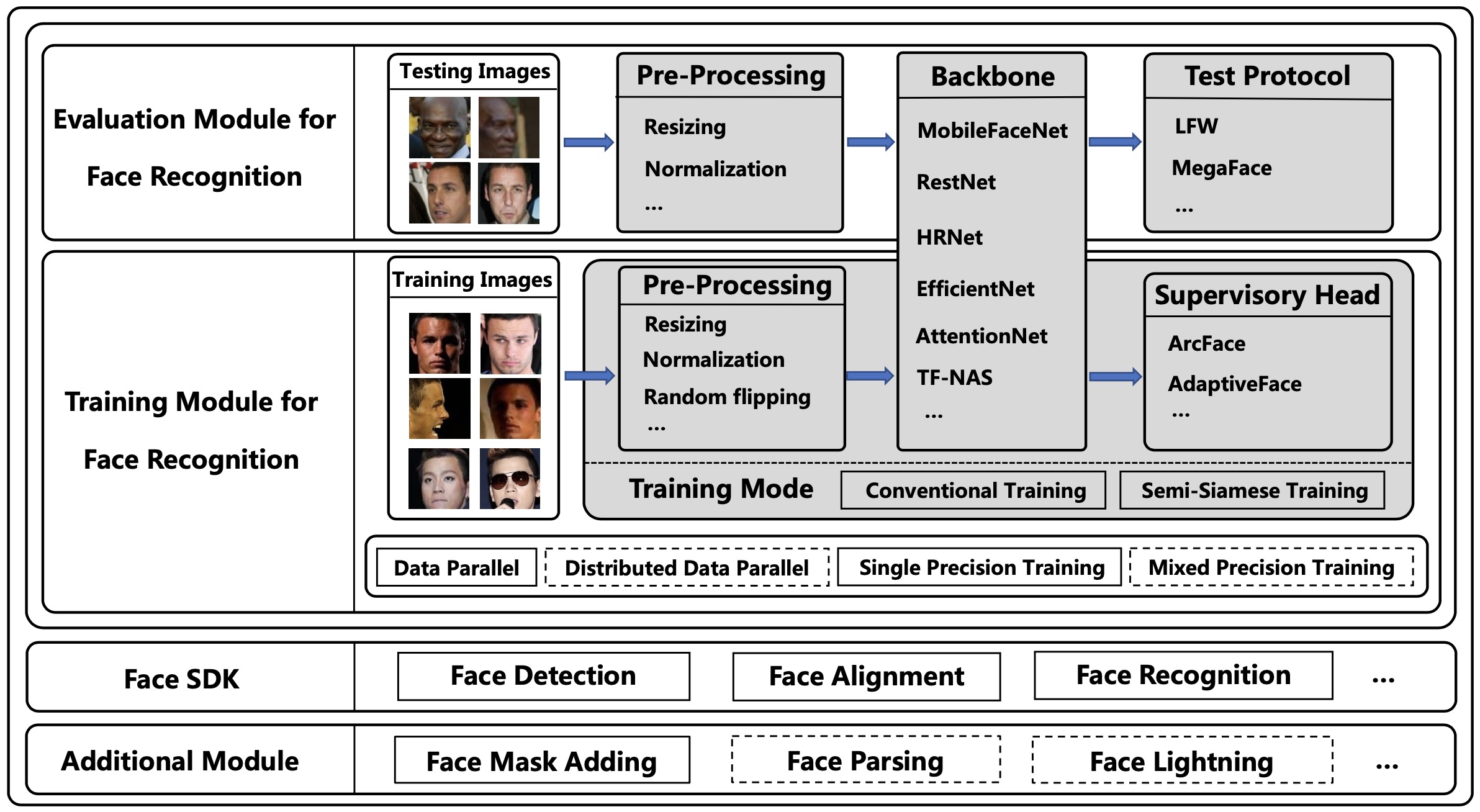}
    \caption{The architecture of FaceX-Zoo. The modules in solid box are those already provided in current version, while the modules in dotted box are to be added in the following versions.}
    \label{fig:architecture}
\end{figure*}

\section{Introduction}
\label{sec:intro}
    Deep learning based face recognition has witnessed great progress in research field.
    Correspondingly, there emerge a number of excellent open-source projects developed for facilitating the experiments and production of deep face recognition networks.
    For example, Facenet~\cite{facenet} is a TensorFlow~\cite{tensorflow2015-whitepaper} implementation of the model proposed by Schroff~\etal~\cite{schroff2015facenet}, which is a classic project for deep face recognition.
    OpenFace~\cite{amos2016openface} is a general face recognition library, especially for the support of mobile device applications.
    InsightFace~\cite{insightface} is a toolbox for 2D\&3D deep face analysis, mainly written in MXNet~\cite{chen2015mxnet}. It includes the commonly-used training data, network settings and loss functions.
    face.evoLVe~\cite{evoLVe} provides a comprehensive face recognition library for face related analytics and applications. Although these projects have been widely used and brought a great deal of convenience, the rapid development of deep face recognition techniques pursuits a significant need of a more comprehensive framework and standard evaluation to facilitate the research and development.
    To this end, we develop a new framework, named FaceX-Zoo, in the form of PyTorch~\cite{paszke2019pytorch} library, which is highly modular, flexible and scalable.
    It is composed of the state-of-the-art training pipeline for discriminative face feature learning, the standard evaluation towards fair comparisons, and the deployment SDK for efficient proof of concept and further applications.
    We release all the source codes and trained models to facilitate the community to develop their own solutions against various real-world challenges from the perspective of training, evaluation, and deployment.
    We hope that FaceX-Zoo is able to provide helpful support to the community and promote the development of face recognition.
    The remaining part of this paper is organized as follows. In Section~\ref{sec:overview}, we depict the structure and the highlight of FaceX-Zoo. In Section~\ref{sec:design_patt}, we introduce the detailed design of this project. Section~\ref{sec:experiment} provides the experiments with respect to the various supervisory heads and backbones that integrated in the training module, and reports the test accuracies on the commonly-used benchmarks which are also provided by the evaluation module. Section~\ref{sec:solution} presents our solutions for two practical situations, \ie shallow face learning and masked face recognition. Finally, we discuss about the future work and give the conclusion in Section~\ref{sec:future} and Section~\ref{sec:conclusion}, respectively.

\section{Overview of FaceX-Zoo}
\label{sec:overview}
\subsection{Architecture}
\label{sec:architecture}
    The overall architecture of FaceX-Zoo is subtly presented in Figure~\ref{fig:architecture}. The whole project mainly consists of four parts: the training module, the evaluation module, the additional module and the face SDK, where the former two modules are the core part of this project. Several components are contained in the training and evaluation modules, including Pre-Processing, Training Mode, Backbone, Supervisory Head and Test Protocol. We elaborate on them as below.\par
    \textbf{Pre-Processing}. This module fulfils the basic transformations on images before sending them to the network. For training, we implement the commonly-used operations, such as resizing, normalization, random cropping, random flipping, random rotation, \etc. One can add the customized operations flexibly, according to various demands. For evaluation, only resizing and normalization are employed. Likewise, the testing augmentations, such as five crops, horizontal flipping, \etc, can also be easily added into our framework by customizing. \par

    \textbf{Training Mode}. The conventional training mode of face recognition is treated as the baseline routine. Concretely, it schedules the training inputs by DataLoader, then sends the inputs to the backbone network for forward passing, and finally computes a criterion as the training loss for backward updating.
    In addition, We consider a practical situation in face recognition that is to train the network with shallow distributed data~\cite{du2020semi}. Accordingly, we integrate a recent training strategy to facilitate the training on shallow face data.
    \par

     \textbf{Backbone}. The backbone network is used to extract the features of face images. We provided a series of state-of-the-art backbone architectures in FaceX-Zoo, which are listed below. Besides, any other architecture choices can be easily customized with the support of PyTorch, as long as modifying the configuration file and adding the architecture definition file.
    \begin{itemize}
    \item MobileFaceNet~\cite{chen2018mobilefacenets}: An efficient network for the applicaiton on mobile devices.
    \item ResNet~\cite{he2016deep}: A series of classic architectures for general vision tasks.
    \item SE-ResNet~\cite{hu2018squeeze}: ResNet equipped with SE blocks that recalibrates the channel wise feature responses.
    \item HRNet~\cite{wang2020deep}: A network for deep high-resolution representation learning.
    \item EfficientNet~\cite{tan2019efficientnet}: A bunch of architectures that scale among depth, width and resolution.
    \item GhostNet~\cite{han2020ghostnet}: A model aiming at generating more feature maps from cheap operations.
    \item AttentionNet~\cite{wang2017residual}: A network built by stacking attention modules to learn attention-aware features.
    \item TF-NAS~\cite{hu2020tfnas}: A series of architectures searched by NAS with the latency constraint.
    \item ResNeSt~\cite{zhang2020resnest}: A series of ResNet-style networks with split-attention blocks.
    \item ReXNet~\cite{han2021rexnet}: A series of models with effective channel configuration and parameterization.
    \item RepVGG~\cite{ding2021repvgg}: A VGG-like architecture realized by structural re-parameterization.
    \item LightCNN~\cite{wu2018lightcnn,wu2020eel}: A light model with max feature map activations for fast face recognition.
    \end{itemize}

    \textbf{Supervisory Head}. Supervisory Head is defined as the supervision single and its corresponding computation module towards accurate face recognition. In order to learn discriminative features for face recognition, the predicted logits are usually processed by some specific operations, such as normalization, scaling, adding margin, \etc, before sending to the softmax layer. We implement a series of softmax-style losses in FaceX-Zoo as follows:
    \par
     \begin{itemize}
    \item AM-Softmax~\cite{wang2018additive}: An additive margin loss that adds a cosine margin penalty to the target logit.
    \item ArcFace~\cite{deng2019arcface}: An additive angular margin loss that adds a margin penalty to the target angle.
    \item AdaCos~\cite{zhang2019adacos}: A cosine-based softmax loss that is hyperparameter-free and adaptive scaling.
    \item AdaM-Softmax~\cite{liu2019adaptiveface}: An adaptive margin loss that can adjust the margins for different classes adaptively.
    \item CircleLoss~\cite{sun2020circle}: A unified formula that learns with class-level labels and pair-wise labels.
    \item CurricularFace~\cite{huang2020curricularface}: An loss function that adaptively adjusts the importance of easy and hard samples during different training stages.
    \item MV-Softmax~\cite{wang2020mis}: A loss function that adaptively emphasizes the mis-classified feature vectors to guide the discriminative feature learning.
    \item NPCFace~\cite{zeng2020npcface}:A loss function that emphasizes the training on both the negative and positive hard cases.
    \end{itemize}

    \textbf{Test protocol}. There are various benchmarks to measure the accuracy of face recognition models. Many of them focus on specific face recognition challenges, such as cross age, cross pose, and cross race. Among them, the commonly used test protocols are mainly based on the benchmarks of LFW~\cite{huang2008labeled} and MegaFace~\cite{kemelmacher2016megaface}.
     We integrates these protocols into FaceX-Zoo with simple usage and clear instruction, by which people can easily test their models on single or multiple benchmarks via simple configurations.
    Besides, it is convenient to extend additional test protocols by adding the test data and parsing the test pairs.
     \textbf{It is worth noting that a masked face recognition benchmark based on MegaFace is provided as well}. \par
    \begin{itemize}
    \item LFW~\cite{huang2008labeled}: It contains 13,233 web-collected images of 5,749 identities with the pose, expression and illumination variations. We report the mean accuracy of 10-fold cross validation on this classic benchmark.
    \item CPLFW~\cite{zheng2018cross}: It contains 11,652 images of 3,930 identities, which focuses on cross-pose face verification. Following the official protocol, the mean accuracy of 10-fold cross validation is adopted.
    \item CALFW~\cite{zheng2017cross}: It contains 12,174 images of 4,025 identities, aiming at cross-age face verification. The mean accuracy of 10-fold cross validation is adopted.
    \item AgeDB30~\cite{moschoglou2017agedb}: It contains 12,240 images of 440 identities, where each test pair has an age gap of 30 years. We report the mean accuracy of 10-fold cross validation.
    \item  RFW~\cite{Wang_2019_ICCV}: It contains 40,607 images of 11,430 identities, which is proposed to measure the potential racial bias in face recognition. There are four test subsets in RFW, named African, Asian, Caucasian and Indian, and we report the mean accuracy of each subset, respectively.
    \item MegaFace~\cite{kemelmacher2016megaface}: It contains 80 probe identities with 1 million gallery distractors, aiming at evaluating large-scale face recognition performance. We report the Rank-K identification accuracy on MegaFace.
    \item MegaFace-Mask: It contains the same probe identities and gallery distractors with MegaFace~\cite{kemelmacher2016megaface}, while each probe image is added by a virtual mask. This protocol is designed to evaluate large-scale masked face recognition performance. More details can be found in Section~\ref{sec:mask}. We report the Rank-K identification accuracy on MegaFace-Mask.
    \end{itemize}

\subsection{Characteristics}
\label{sec:char}
    \textbf{Modular and extensible design}.
    As described above, FaceX-Zoo is designed to be modular and extensible.
    It consists of a set of modules with respective functions. Most of the modules are developed following the principle of object-oriented design, so that the scalability is highly promoted.
    One can easily add new training modes, backbones, supervisory heads and data samplers to the training module, as well as more test protocols to the evaluation module.
    Last but not the least, we provide the face SDK and the additional module for the efficient deployment and flexible extension according to various demands.
    \par
    \textbf{State-of-the-art training}.
    We provide several state-of-the-art practices for face recognition model training, such as the complete pre-processing operations for data augmentation, the various backbone networks for model ensemble, the softmax-style loss functions for discriminative feature learning, and the Semi-Siamese Training mode for practical shallow face learning.
    \par
    \textbf{Easy to use and deploy}.
    We release all the codes, models and training logs for reproducing the state-of-the-art results. Besides, we provide a simple yet fully functional face SDK written in python, which acts as a demo to help the users learn the usage of each module and develop the further versions. \par
    \textbf{Standardized evaluation module}. The commonly used evaluation benchmarks for face recognition are in need of a unified implementation for efficient and fair evaluation.
    For example, the official test protocol of MegaFace is implemented as a bin file, leading to inconvenient application in many evaluation conditions.
    FaceX-Zoo provides a standard and open-source implementation for evaluating on LFW-based and MegaFace-based benchmarks. Users can evaluate models on various benchmarks by editing the configuration file efficiently.
    We will also release the 106-point facial landmarks defined as~\cite{liu2019grand} so that users can utilize them for face alignment of these benchmarks.

    \par
    \textbf{Support for masked face recognition}. Recently, due to the pandemic of COVID-19, masked face recognition has attracted increasing attention.
    In order to develop such a model, three essential components are indispensable: masked face training data, masked face training algorithm and masked face evaluation benchmark. FaceX-Zoo is the very project that provides all the three components via the 3D face mask adding technique.

\section{Detailed Design}
\label{sec:design_patt}
In this section, we describe the design of the training module (Figure~\ref{fig:training_module}), the evaluation module (Figure~\ref{fig:test_module}), and the face SDK (Figure~\ref{fig:sdk_module}) in details, which are modular  and extensible.

\begin{table*}
\renewcommand{\arraystretch}{1.2}
\caption{The performance (\%) with different backbones, where RFW (Afr), RFW (Asi), RFW (Cau) and RFW (Ind) denote the African, Asian, Caucasian and Indian test protocols in RFW, respectively. Apart from MegaFace, we report the mean accuracies on these benchmarks. For MegaFace, we report the Rank-1 accuracy.}
\label{tab:exp_backbone}
\small
\centering
\begin{tabular}{lccccccccc}
  \hline
  Backbone & LFW & CPLFW & CALFW & AgeDB & RFW (Afr) & RFW (Asi) & RFW (Cau) & RFW (Ind) & MegaFace\\
  \hline
  \hline
  MobileFaceNet & 99.57 & 83.33 & 93.82 & 95.97 & 88.73 & 88.02 & 95.70 & 90.85 & 90.39 \\
 ResNet50-ir & 99.78 & 88.20 & 95.47 & 97.77 & 95.25 & 93.95 & 98.57 & 95.80 & 96.67 \\
 ResNet152-irse & 99.85 & 89.72 & 95.56 & 98.13 & 95.85 & 95.43 & 99.08 & 96.27 & 97.48 \\
 HRNet & 99.80 & 88.89 & 95.48 & 97.82 & 95.87 & 94.77 & 99.08 & 95.93 & 97.32 \\
 EfficientNet-B0 & 99.55 & 84.72 & 94.37 & 96.63 & 89.67 & 89.32 & 96.10 & 91.93 & 91.38 \\
 TF-NAS-A & 99.75 & 85.90 & 94.87 & 97.23 & 91.97 & 91.62 & 97.43 & 93.33 & 94.42 \\
 LightCNN-29 & 99.57 & 82.60 & 93.87 & 95.78 & 88.32 & 87.83 & 95.33 & 90.88 & 89.32 \\
 GhostNet & 99.65 & 85.30 & 93.92 & 96.08 & 88.67 & 88.48 & 95.13 & 90.63 & 87.88 \\
 Attention-56 & 99.88 & 89.18 & 95.65 & 98.12 & 96.52 & 95.72 & 99.13 & 96.83 & 97.75 \\
 ResNeSt50 & 99.80 & 89.98 & 95.55 & 97.98 & 95.45 & 94.52 & 98.65 & 95.57  & 97.08 \\
 ReXNet\_1.0 & 99.65 & 84.68 & 94.58 & 96.70 & 90.73 & 89.95 & 96.58 & 92.17  & 93.17 \\
 RepVGG\_B1 & 99.82 & 87.55 & 95.50 & 97.78 & 94.57 & 93.88 & 98.73 & 95.03  & 96.74 \\
  \hline
\end{tabular}
\end{table*}
\begin{table*}
\renewcommand{\arraystretch}{1.2}
\caption{The performance (\%) with different supervisory heads, where RFW (Afr), RFW (Asi), RFW (Cau) and RFW (Ind) denote the African, Asian, Caucasian and Indian test protocols in RFW, respectively. Apart from MegaFace, we report the mean accuracies on these benchmarks. For MegaFace, we report the Rank-1 accuracy.}
\label{tab:exp_head}
\small
\centering
\begin{tabular}{lccccccccc}
  \hline
  Supervisory head & LFW & CPLFW & CALFW & AgeDB & RFW (Afr) & RFW (Asi) & RFW (Cau) & RFW (Ind) & MegaFace\\
  \hline
  \hline
  AM-Softmax & 99.58 & 83.63 & 93.93 & 95.85 & 88.38 & 87.88 & 95.55 & 91.18 & 88.92 \\
 AdaM-Softmax & 99.58 & 83.85 & 93.50 & 96.02 & 87.90 & 88.37 & 95.32 & 91.13 & 89.40 \\
 AdaCos & 99.65 & 83.27 & 92.63 & 95.38 & 85.88 & 85.50 & 94.35 & 88.27 & 82.95 \\
 ArcFace & 99.57 & 83.68 & 93.98 & 96.23 & 88.22 & 88.00 & 95.13 & 90.70 & 88.39 \\
 MV-Softmax & 99.57 & 83.33 & 93.82 & 95.97 & 88.73 & 88.02 & 95.70 & 90.85 & 90.39 \\
 CurricularFace & 99.60 & 83.03 & 93.75 & 95.82 & 88.20 & 87.33 & 95.27 & 90.57 & 87.27 \\
  CircleLoss & 99.57 & 83.42 & 94.00 & 95.73 & 89.25 & 88.27 & 95.32 & 91.48 & 88.75 \\
  NPCFace & 99.55 & 83.80 & 94.13 & 95.87 & 88.08 & 88.20 & 95.47 & 91.03 & 89.13 \\
  \hline
\end{tabular}
\end{table*}

\subsection{Training Module}
\label{sec:training_module}
\begin{figure}[t]
    \centering
    \includegraphics[width=\linewidth]{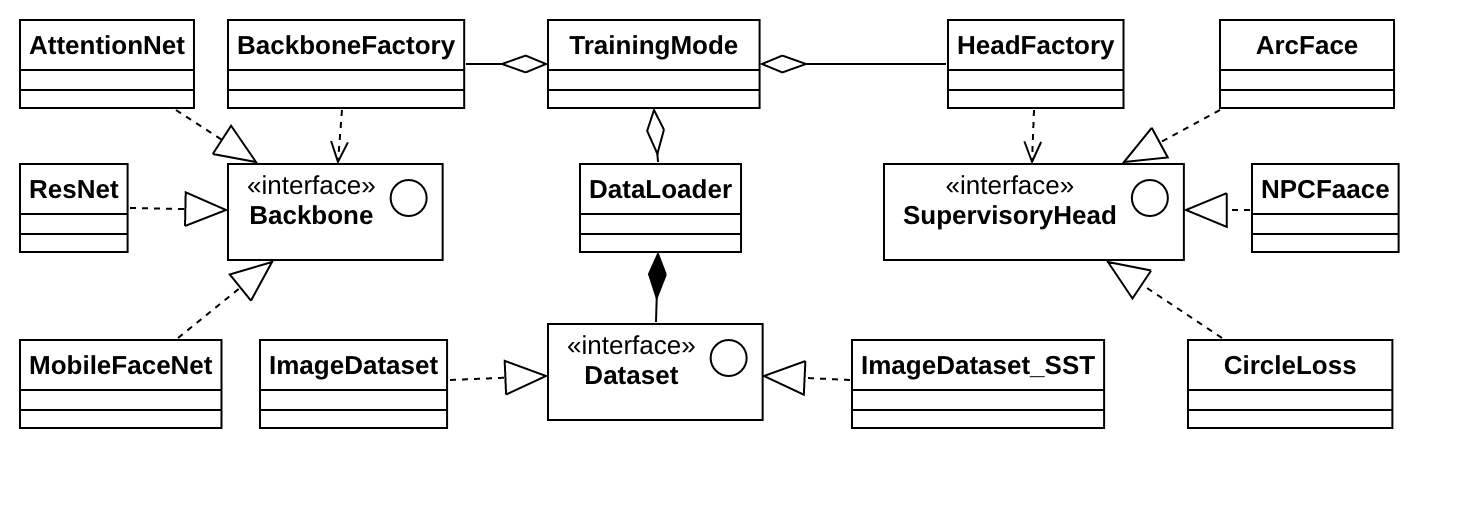}
    \caption{The class diagram of training module.}
    \label{fig:training_module}
\end{figure}

As shown in Figure~\ref{fig:training_module}, the TrainingMode is the core class to aggregate all the other classes in the training module. There are mainly three classes aggregated in the TrainingMode: (1) BackboneFactory is a factory class to provide the backbone network; (2) HeadFactory is a factory class to produce the supervisory head according to the configuration; (3) DataLoader is in charge of loading the training data.

\subsection{Evaluation Module}
\label{sec:test_module}
\begin{figure}[t]
    \centering
    \includegraphics[width=\linewidth]{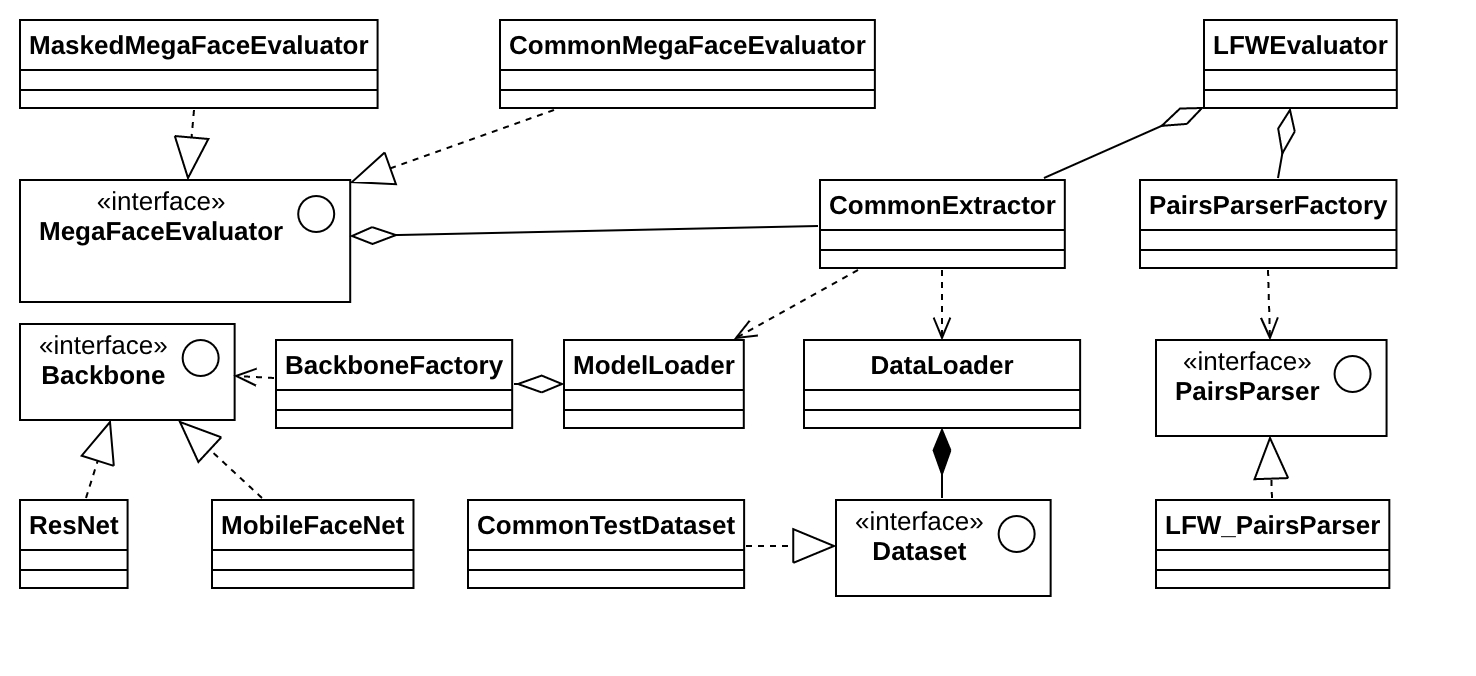}
    \caption{The class diagram of evaluation module.}
    \label{fig:test_module}
\end{figure}

As depicted in Figure~\ref{fig:test_module}, the LFWEvaluator and the MegaFaceEvaluator are the core classes in the evaluation module. Both of them contain the class of CommonExtrator for face feature extraction. The CommonExtrator class depends on the ModelLoader class and the DataLoader class, where the former loads the models and the later loads the test data. Besides, the LFWEvaluator class also aggregates the PairsParseFactory class for parsing the test pairs in each test set. Differently, we split two classes for MegaFace-based evaluations, named CommonMegaFaceEvaluator and MaskedMegafaceEvaluator, for the MegaFace evaluation and the MegaFace-Mask evaluation, respectively. Both of them are inherited from the MegaFaceEvaluator class.

\begin{figure}[t]
    \centering
    \includegraphics[width=\linewidth]{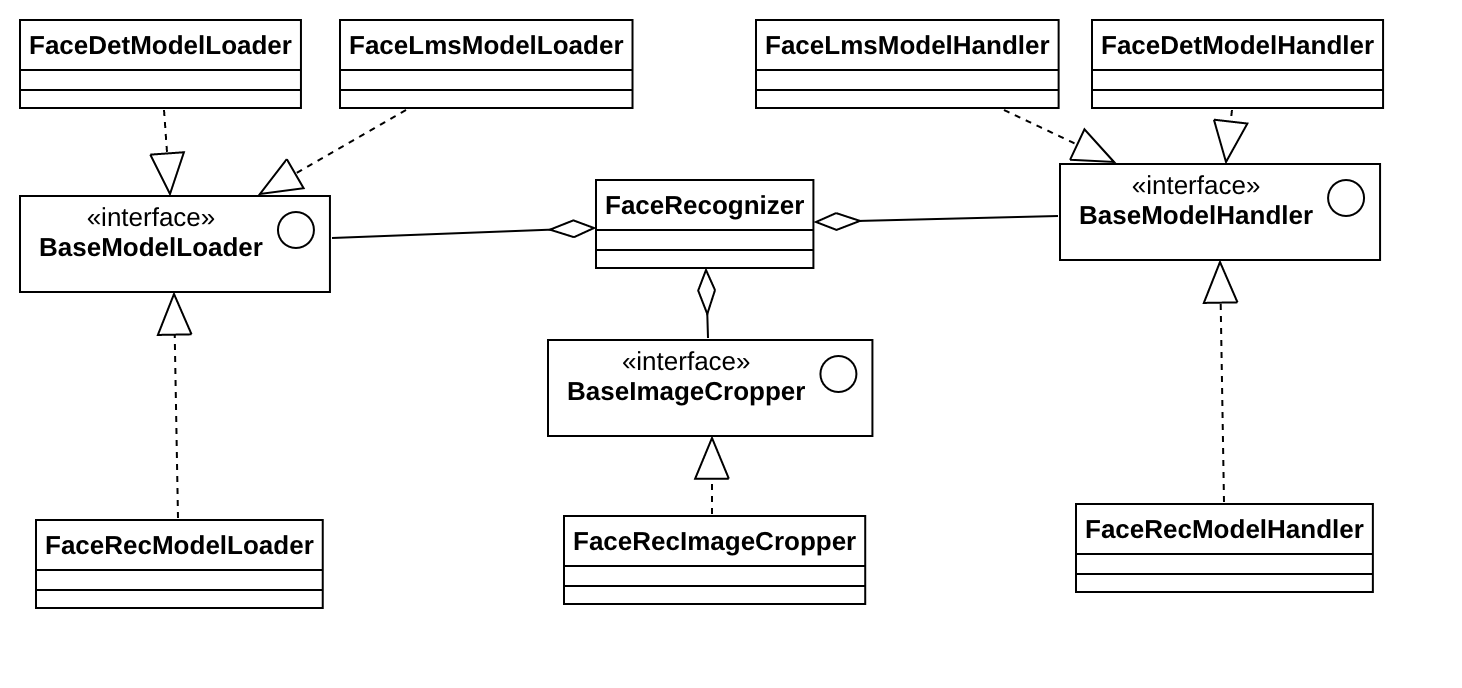}
    \caption{The class diagram of the face SDK.}
    \label{fig:sdk_module}
\end{figure}
\subsection{Face SDK}
\label{sec:sdk}
In order to validate and demonstrate the effectiveness of the trained models for face recognition in a convenient way, we provide a simple yet fully functional module of Face SDK. As shown in Figure~\ref{fig:sdk_module}, Face SDK includes three core classes, named ModelLoader, ImageCropper and ModelHandler. The ModelLoader class is used to load the models of face detection, face landmark localization and face feature extraction. The ImageCropper class is used to crop the facial area from the input image according to the detected facial landmarks, and output the normalized face crop. The ModelHandler class provides pre-processing and post-processing operations, as well as the inference interface.\par

In Face SDK, we provide a series of models, \ie face detection, facial landmark localization, and face recognition, for the non-masked face recognition and masked face recognition scenarios. Specifically, for the non-masked face recognition scenario, we train the face detection model by RetinaFace~\cite{deng2019retinaface} on the WiderFace dataset~\cite{yang2016wider}. The facial landmark localization model is trained by PFLD~\cite{guo2019pfld} on the JD-landmark dataset~\cite{liu2019grand}. We train the face recognition model with MobileFaceNet~\cite{chen2018mobilefacenets} and MV-Softmax~\cite{wang2019mis} on MS-Celeb-1M-v1c~\cite{trillionpairs}. For the masked face recognition scenario, we train the models with the same algorithms as the non-masked scenario while the training data is expanded by our FMA-3D method described in Section~\ref{sec:mask}. We will continuously update the models with more methods in the future.

\section{Experiments of SOTA Components}
\label{sec:experiment}
To facilitate the readers to reproduce and fulfil their own works with our framework, we conduct extensive experiments about the backbone and supervisory head with the state-of-the-art methods.
The adopted backbones and supervisory heads are listed in Section~\ref{sec:architecture}.
We use MS-Celeb-1M-v1c~\cite{trillionpairs} as the training data, which is well cleaned. 
For clear presentation, in the experiments of backbone, we adopt the same supervisory head, \ie MV-Softmx~\cite{wang2020mis}; in the experiments of supervisory head, we adopt the same backbone, \ie MobileFaceNet~\cite{chen2018mobilefacenets}.
The remaining settings are kept the same for each trial.
Four NVIDIA Tesla P40 GPUs are employed for training.
We set the total epoch to 18 and the batch size to 512 for training. The learning rate is initialized as 0.1, and divided by ten at the epoch 10, 13 and 16.
The test results of the experiments of backbone and supervisory head are shown in Table~\ref{tab:exp_backbone} and Table~\ref{tab:exp_head}, respectively. One can refer to these results for guiding and verifying the usage of our framework.


\section{Task-specific Solutions}
\label{sec:solution}
In this section, we present to use the specific solutions for handling two challenging tasks of face recognition within the framework of FaceX-Zoo, including
Semi-Siamese Training~\cite{du2020semi} for shallow face learning, and the masked face recognition for the recent demand caused by the pandemic of COVID-19.

\begin{table*}
\renewcommand{\arraystretch}{1.2}
\caption{The performance (\%) of different training modes applied on shallow data.
RFW (Afr), RFW (Asi), RFW (Cau) and RFW (Ind) denote the African, Asian, Caucasian and Indian test protocols in RFW, respectively.
}
\label{tab:exp_shallow}
\small
\centering
\begin{tabular}{lcccccccc}
  \hline
  Training Mode & LFW & CPLFW & CALFW & AgeDB & RFW (Afr) & RFW (Asi) & RFW (Cau) & RFW (Ind)\\
  \hline
  \hline
  Conventional Training & 91.77 & 61.56 & 76.52 & 73.90 & 61.35 & 67.38 & 73.27 & 70.12  \\
  Semi-siamese Training & 99.38 & 82.53 & 91.78 & 93.60 & 85.03 & 85.25 & 92.80 & 87.40  \\
  \hline
\end{tabular}
\end{table*}
\subsection{Shallow Face Learning}
\label{sec:shallow}
\textbf{Background}. In many real-world scenarios of face recognition, the training dataset is limited in depth, \eg only two face images are available for each ID.
This task, which is so called Shallow Face Learning as described in \cite{du2020semi}, is problematic to the conventional training methods for face recognition.
The shallow face data severely lacks the intra-class diversity for each ID, and leads to the collapse of feature dimension against effective training.
Consequently, the trained network suffers from either model degeneration or over-fitting.
As suggested in \cite{du2020semi}, we adopt Semi-Siamese Training (SST) to tackle this issue. Furthermore, we implement it by the framework of FaceX-Zoo, in which the upstream and downstream stages (\ie efficient data reading and unified automatic evaluation) complete the pipeline and facilitate the users to employ SST for model production.

\textbf{Experiments and results}.
For a quick verification of the effectiveness of FaceX-Zoo towards shallow face learning, we employ an off-the-shelf architecture, \ie MobileFaceNet, as the model backbone, and perform a comparison experiment between the conventional training and SST.
Following the settings of \cite{du2020semi}, the training dataset is constructed by randomly selecting two facial images from each ID of MS-Celeb-1M-v1c, called MS-Celeb-1M-v1c-Shallow.
The training epoch is set to 250 and the batch size is set to 512. The learning rate is initialized as 0.1, and divided by ten at the epoch 150, 200, 230.
The test results on LFW, CPLFW, CALFW, AgeDB and RFW are presented in Table~\ref{tab:exp_shallow}, which verifies the effectiveness of poly-mode training on the shallow data.

\subsection{Masked Face Recognition}
\label{sec:mask}
\textbf{Background}.  Due to the recent world-wide COVID-19 pandemic, masked face recognition has become a crucial application demand in many scenarios. However, few masked face datasets are available for training and evaluation.
To address this issue, we empower the framework of FaceX-Zoo to add virtual mask to the existing face images by the specialized module, named FMA-3D (3D-based Face Mask Adding). \par
\begin{figure}[h]
    \centering
    \includegraphics[width=\linewidth]{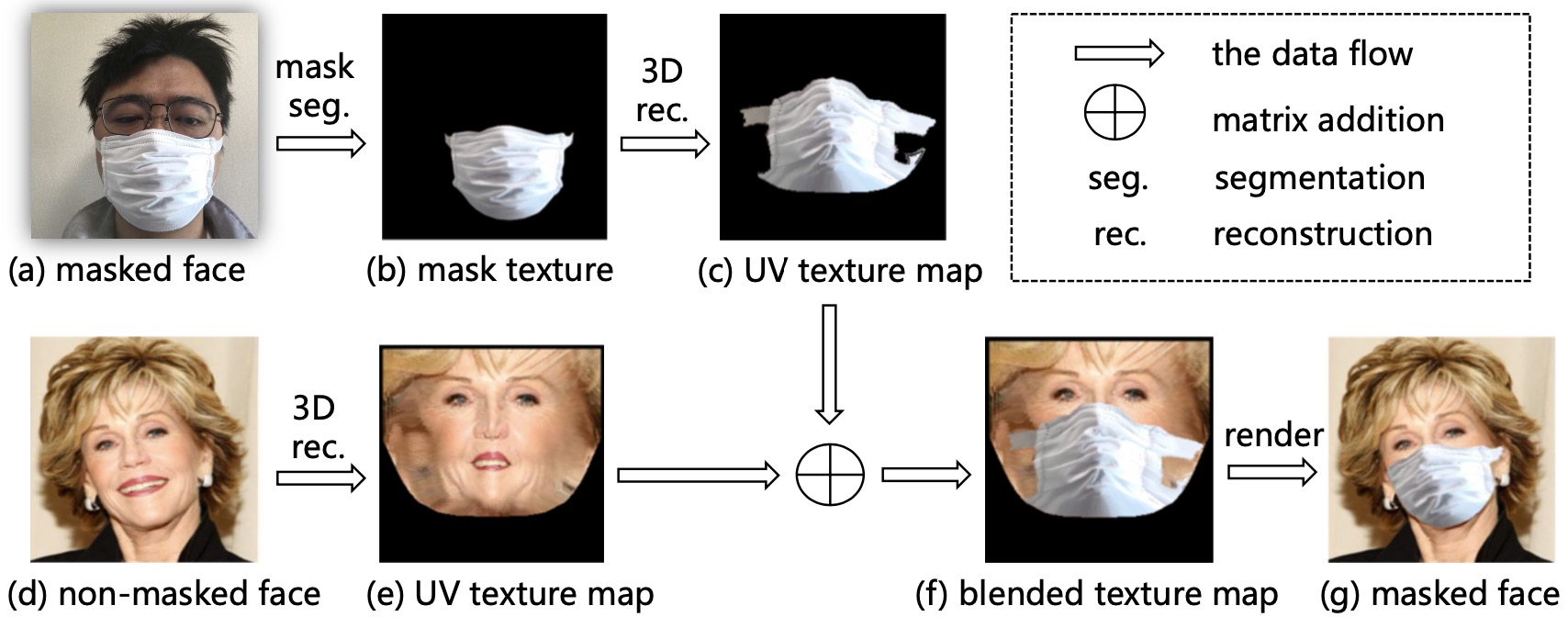}
    \caption{The method for wearing virtual masks on face image.
    The mask template can be sampled from various choices subject to the input masked face.}
    \label{fig:mask}
\end{figure}
\begin{figure}[h]
    \centering
    \includegraphics[width=\linewidth]{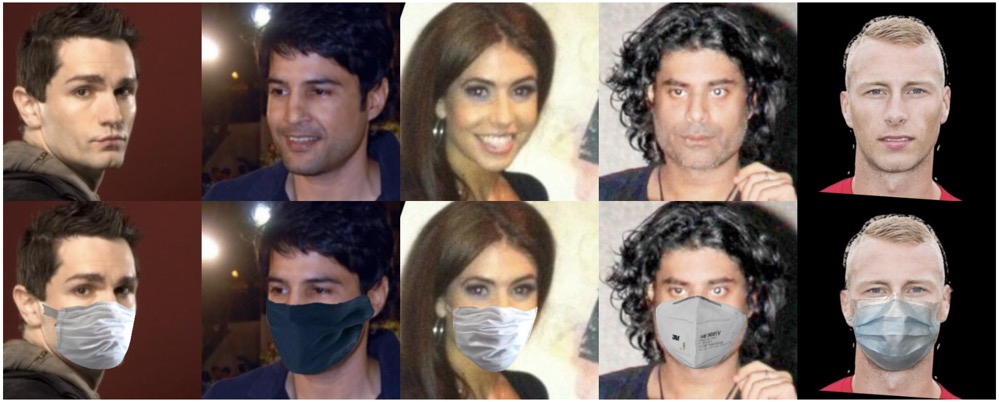}
    \caption{
    Top: the original non-masked face images.
    Bottom: the masked face image synthesized by FMA-3D.}
    \label{fig:mask-sample}
\end{figure}

\textbf{FMA-3D}.
Given a real masked face image $A$ (Fig.~\ref{fig:mask}(a)) and a non-masked face image $B$ (Fig.~\ref{fig:mask}(d)), we synthesize a photo-realistic masked face image with the mask from $A$ and the facial area from $B$.
First, we utilize a mask segmentation model~\cite{liu2020new} to extract the mask area from image $A$ (Fig.~\ref{fig:mask}(b)), and then map the texture map into UV space by the 3D face reconstruction method PRNet~\cite{feng2018joint} (Fig.~\ref{fig:mask}(c)). For image $B$, we compute the texture map in UV space in the same way of A (Fig.~\ref{fig:mask}(e)). Next, we blend the mask texture map and the face texture map in UV space as Fig.~\ref{fig:mask}(f) shows. Finally, the masked face image is synthesized (Fig.~\ref{fig:mask}(g)) by rendering the blended texture map according to the UV position map of image $B$. Fig.~\ref{fig:mask-sample} shows more cases of masked face image synthesized by FMA-3D.

Compared with the 2D-based and GAN-based methods, our method shows superior performance on the robustness and fidelity, especially for the large head poses.

\textbf{Training masked face recognition model.}
Resorting to our FMA-3D, it is convenient to synthesize large number of masked face images from the existing non-masked datasets, such as MS-Celeb-1M-v1c. Since the existing datasets already have the ID annotation, we can directly employ them for training the face recognition network without additional labeling.
The training method can be either the conventional routine or SST, as well as the training head and backbone can be instantiated with the choices integrated in FaceX-Zoo.
Note that the testing benchmark can be augmented from non-masked to masked version in the same manner.

\begin{figure}[h]
    \centering
    \includegraphics[width=\linewidth]{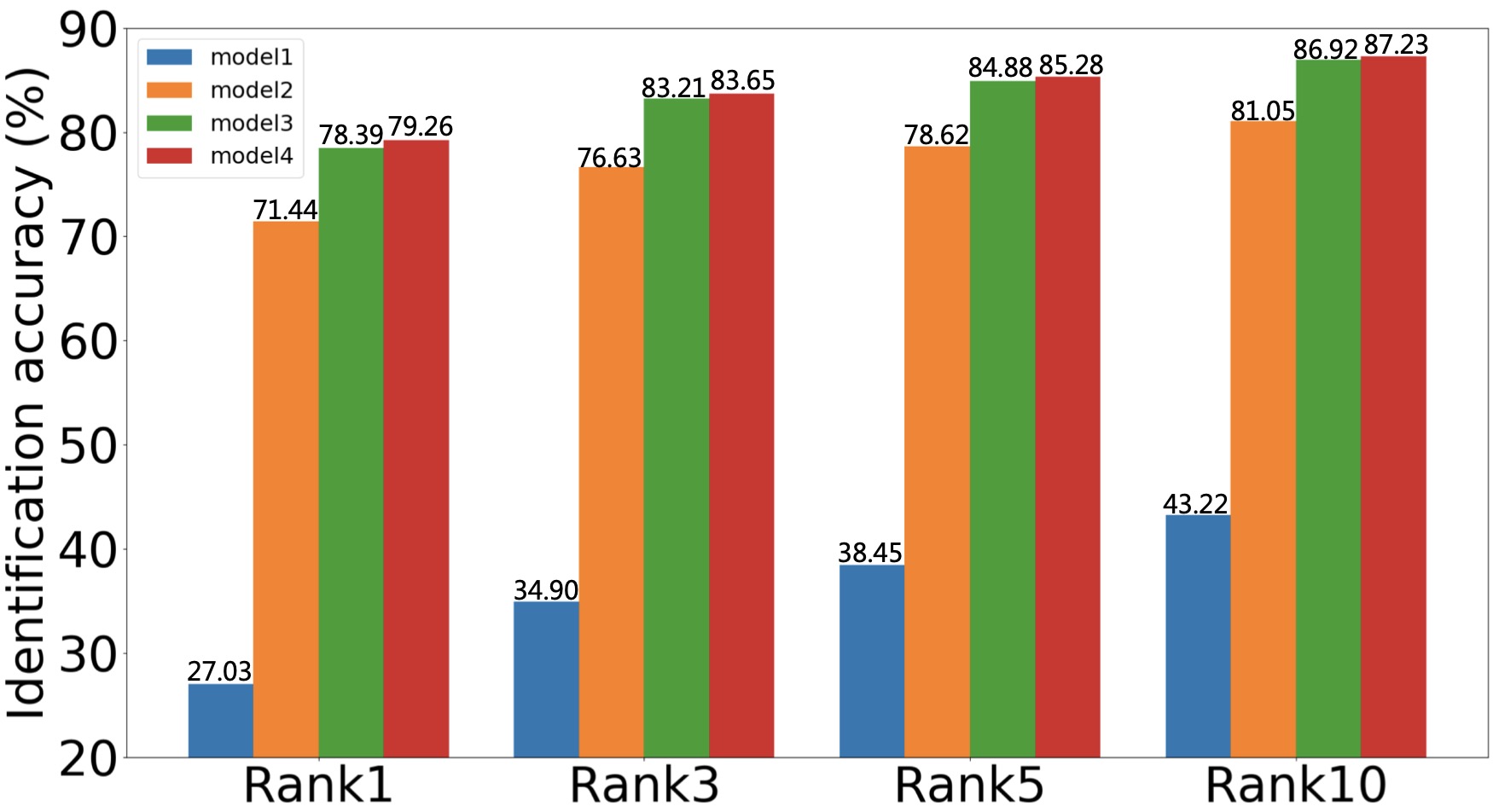}
    \caption{Rank-K identification accuracy on MegaFace-Mask. Zoom in for better view.}
    \label{fig:mask-bar}
\end{figure}

\textbf{Experiments and results.}
By using FMA-3D, we synthesize the training data from MS-Celeb1M-v1c to its masked version, named MS-Celeb1M-v1c-Mask. It includes the original face images of each identity in MS-Celeb1M-v1c, as well as the masked face images corresponding to the original ones.
We choose MobileFaceNet as the backbone, and MV-Softmax as the supervisory head. The model is trained for 18 epochs with a batch size of 512. The learning rate is initialized as $0.1$, and divided by ten at the epoch 10, 13 and 16.
To evaluate the model on masked face recognition task, we synthesize the masked facial datasets based on MegaFace by using FMA-3D, named MegaFace-mask, which contains the masked probe images and remains the gallery images non-masked.
As shown in Figure~\ref{fig:mask-bar}, we conduct comparison experiments among four scenarios. Specifically, $model1$ is the baseline which is trained on MS-Celeb1M-v1c; $model2$ is also trained on MS-Celeb1M-v1c, but only the upper half of face is cropped for training, which can be regarded as a naive manner to eliminate the adverse effect of mask; $model3$ is trained on MS-Celeb1M-v1c-Mask; $model4$ is the ensemble of $model2$ and $model3$.
We can see that the rank1 accuracy of baseline model is 27.03\%. By only utilizing the upper half of face, the performance of $model2$ is improved to 71.44\%. $Model3$ achieves the best performance of 78.39\% in single models with the help of synthesized masked face images. By combining $model2$ and $model3$, the rank1 accuracy is further improved to 79.26\%.

\par



\section{Future Work}
\label{sec:future}
In the future, we will try to improve FaceX-Zoo from three aspects: breadth, depth, and efficiency.
First, more additional modules will be included, such as face parsing and face lightning, to thereby enrich the functionality ``\textit{X}'' in FaceX-Zoo.
Second, the modules of backbone architecture and supervisory heads will be continually supplemented along with the development of deep learning techniques.
Third, we will try to improve the training efficiency via distributed data parallel technique and mixed precision training.

\section{Conclusion}
\label{sec:conclusion}
In this work, we introduce a highly modular and scalable open-source framework for face recognition, namely FaceX-Zoo.
It is easy to install and utilize.
The Training Module enable users to train face recognition networks with various choices of backbone and supervisory head.
The Training Mode includes both the conventional routine and the specific solution for shallow face learning.
The Evaluation Module provides an automatic evaluation benchmark for standard and convenient testing.
Face SDK provides modules for the whole pipeline, \ie face detection, face landmark localization, and face feature extraction, for face recognition. It can be taken as a baseline as well as further development towards deployment.
Besides, the Additional Module supports training and testing on masked face recognition via 3D virtual mask adding technique.

All the source codes are released along with the logs and trained models.
One can easily play with this framework as a prototype, and develop his own work from this baseline.



{\small
	\bibliographystyle{ieee_fullname}
	\bibliography{ref}
}
	
\end{document}